\documentclass{article}

\usepackage{PRIMEarxiv}

\usepackage[utf8]{inputenc} 
\usepackage[T1]{fontenc}    
\usepackage[pagebackref,breaklinks,colorlinks]{hyperref}
\usepackage{url}            
\usepackage{booktabs}       
\usepackage{amsfonts}       
\usepackage{nicefrac}       
\usepackage{microtype}      
\usepackage{fancyhdr}       
\usepackage{graphicx}       
\usepackage{amsmath}
\usepackage{amssymb}
\usepackage{bm}
\usepackage{multirow}

\graphicspath{{media/}}     

\title{MoE-based Feature Adapter for Prompt-free Binary Coronary Artery Segmentation in X-ray Angiography
}

\author{
  Lin Xi$^{1,2}$,~~Yingliang Ma$^{2,3,}$\thanks{Corresponding author}\\
  $^{1}$University College London, United Kingdom\\
  $^{2}$University of East Anglia, United Kingdom\\
  $^{3}$King's College London, United Kingdom\\
  \texttt{xilin.chibchin@outlook.com},~~\texttt{yingliang.ma@uea.ac.uk}\\
}

\begin{document}
\maketitle

\begin{abstract}
Accurate segmentation of coronary arteries in X-ray angiography videos is essential for quantitative coronary analysis and image-guided interventions. However, accurate segmentation remains challenging because coronary vessels are thin and exhibit low contrast, while the presence of catheters, guidewires, and complex anatomical background structures can further interfere with vessel delineation. Existing U-Net- and Transformer-based models provide strong baselines, but their shared feature-adaptation pathways may be insufficient for heterogeneous angiographic appearances. In this paper, we propose a prompt-free mixture-of-experts (MoE) feature adapter for binary coronary artery segmentation. Built upon parameter-efficient Vision Transformer adapters, the proposed method uses multiple lightweight experts with input-dependent top-$k$ routing to adaptively refine vessel-related features while limiting active computational cost. Experiments on MOSXAV and external evaluation on XACV show that the proposed method outperforms representative baselines and improves cross-dataset generalisation. These results suggest that MoE-based adapter learning is effective for robust coronary artery segmentation in X-ray angiography videos.
\end{abstract}

\section{Introduction}

X-ray coronary angiography is widely used for the diagnosis and treatment of coronary artery disease. It provides real-time visualisation of coronary arteries during clinical procedures and remains one of the most important imaging modalities in interventional cardiology. Therefore, automatic coronary artery segmentation from angiography videos is an important step for quantitative vessel analysis, treatment planning, image-guided interventions, and downstream computer-assisted diagnosis.

Despite recent advances in medical image segmentation, robust coronary artery segmentation in X-ray angiography remains challenging. Coronary arteries are thin, tortuous, and curvilinear, and they often appear partially discontinuous due to uneven contrast-agent filling and weak distal vessel visibility. These characteristics are consistent with the broader challenge of curvilinear structure segmentation, where thin tubular structures must be delineated under low contrast, uneven appearance, and complex background interference~\cite{mou2021cs2net}. Such difficulties are further compounded by a low signal-to-noise ratio and confounding anatomical or device-related structures, both of which can substantially impair vessel delineation~\cite{gao2022vessel,xi2025catheter}. In particular, catheters and guidewires may exhibit grey-level intensities similar to those of coronary vessels, while motion artefacts can blur vessel boundaries and introduce structural ambiguity. Together, these factors produce strong foreground--background ambiguity. Consequently, robust coronary artery segmentation requires effective modelling of fine curvilinear structures under low-contrast, discontinuous, and cluttered imaging conditions.

Encoder-decoder networks, such as U-Net~\cite{unet} and Attention U-Net~\cite{attentionunet}, have been widely used for medical image segmentation because they combine hierarchical feature extraction with spatial detail recovery. nnU-Net~\cite{nnunet} further provides a strong self-configuring baseline by automatically adapting preprocessing, network design, and training strategies to a given dataset. More recently, Transformer-based segmentation models have shown strong potential for modelling long-range dependencies in medical images~\cite{transunet,swinunet}. For X-ray angiography, recent studies have explored semi-supervised segmentation with noisy pseudo-labels and introduced MOSXAV as a benchmark for angiography video segmentation~\cite{xi2026diffusion}, while few-shot video object segmentation has been investigated using local matching and spatio-temporal consistency~\cite{xi2026fsvos}. However, robust binary coronary artery segmentation remains difficult because vessel and background regions exhibit highly heterogeneous visual patterns. Thin distal branches require the preservation of fine local details, weakly opacified vessels demand sensitivity to subtle intensity variations, and regions overlapping with catheters or guidewires require effective suppression of non-vessel responses. A single shared feature transformation may therefore lead to missed small branches, over-smoothed boundaries, or false positives. This motivates a more flexible feature adaptation mechanism that can refine vessel-related representations according to local appearance variations while maintaining robust global semantic understanding.

Parameter-efficient adaptation provides a promising alternative for adapting large visual backbones to downstream medical segmentation tasks~\cite{houlsby2019parameter,lora,vpt,vitadapter,medsa,samed}. AdaptFormer~\cite{adaptformer} introduces lightweight adapter modules for Vision Transformers and demonstrates that compact trainable branches can effectively adapt transformer representations with limited additional parameters. Nevertheless, the original adapter design relies on a single adapter branch, which applies the same feature transformation to all input representations. For coronary artery segmentation, such a uniform adaptation strategy may be insufficient to handle the heterogeneous angiographic appearances caused by thin vessels, low contrast, contrast-filled arterial segments, catheter overlap, and complex anatomical background structures~\cite{mou2021cs2net,gao2022vessel,xi2025catheter}. This motivates an input-dependent feature adaptation strategy, where multiple expert branches can refine vessel-related representations according to local appearance variations, inspired by mixture-of-experts and conditional computation~\cite{jacobs1991adaptive,shazeer2017outrageously,fedus2022switch}.

In this work, we propose a prompt-free mixture-of-experts feature-adapter framework for binary coronary artery segmentation in X-ray angiography videos. Instead of using a single AdaptFormer-style adapter, the proposed framework introduces multiple lightweight adapter experts and an input-dependent routing mechanism to dynamically combine the most relevant experts for each feature representation. Inspired by sparse mixture-of-experts learning~\cite{shazeer2017moe}, this design enables adaptive feature refinement for heterogeneous coronary angiography patterns while maintaining computational efficiency through sparse top-$k$ routing. Since the method is prompt-free, it does not require manual prompts or additional prompt engineering during training or inference.

The main contributions of this paper are summarised as follows:
\begin{itemize}
\item We propose a MoE-based AdaptFormer-style lightweight adapters for input-dependent feature adaptation in coronary angiography segmentation.
\item We introduce sparse top-$k$ expert routing to improve adaptation flexibility while limiting active computational cost.
\item We evaluate the proposed method on MOSXAV and XACV, demonstrating improved segmentation performance and cross-dataset generalisation over representative baselines.
\end{itemize}


\section{Method}

\subsection{Problem Formulation}

Given an X-ray angiography frame $\bm{I} \in \mathbb{R}^{H \times W}$, the goal is to predict a binary coronary artery segmentation mask $\bm{Y} \in \{0,1\}^{H \times W}$, where vessel pixels are labelled as foreground and all non-vessel pixels are labelled as background. The predicted probability map is denoted as $\hat{\bm{Y}} \in [0,1]^{H \times W}$.

The proposed model is formulated as:
\begin{equation}
    \hat{\bm{Y}} = D(E(\bm{I}; \theta_E, \theta_A); \theta_D),
\end{equation}
where $E$ is the transformer-based encoder, $D$ is the segmentation decoder, $\theta_E$ and $\theta_D$ denote the backbone parameters, and $\theta_A$ denotes the parameters of the proposed MoE-based feature adapter modules.

\subsection{Overview of MoE-based Feature Adapter}

\begin{figure}[htbp]
    \centering
    \includegraphics[width=0.9\linewidth]{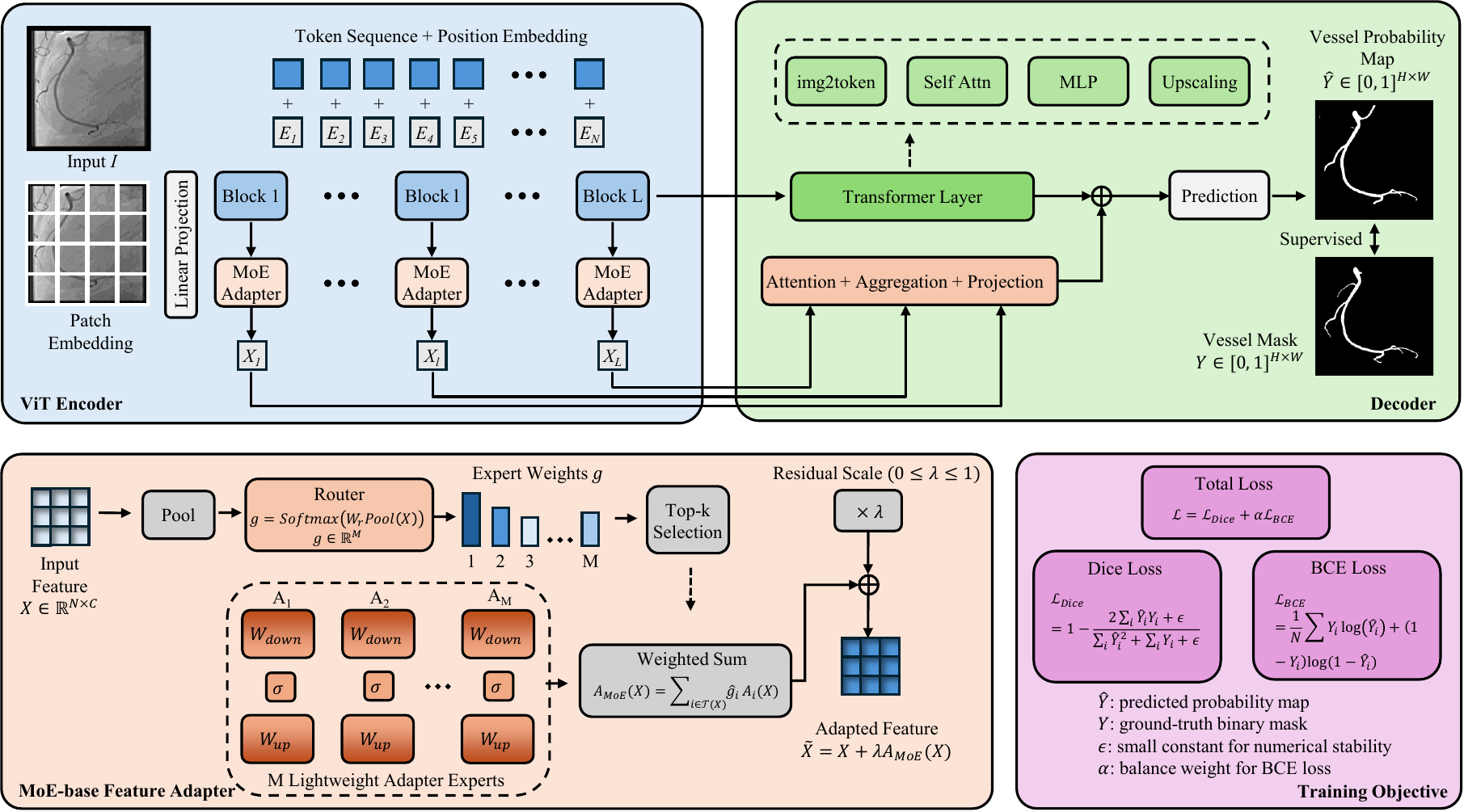}
    \caption{Overall architecture of the proposed network.}
    \label{fig:framework}
\end{figure}

MoE-based Feature Adapter follows an encoder-decoder segmentation design. The encoder extracts hierarchical feature representations from the input angiography frame. At selected transformer blocks, we insert MoE adapter modules to refine intermediate token features. The decoder progressively fuses multi-scale features and predicts the final binary vessel probability map.

The key component is the MoE-based feature adapter. Instead of using a single adapter branch, the proposed module contains multiple lightweight adapter experts and a router. For each input feature, the router predicts the importance of each expert. Only the top-$k$ experts are activated, and their outputs are combined to generate an adaptive residual update.

\subsection{AdaptFormer-style Lightweight Adapter}

Given an intermediate token feature $\bm{X} \in \mathbb{R}^{N \times C}$, where $N$ is the number of tokens and $C$ is the channel dimension, a lightweight adapter performs bottleneck feature transformation:
\begin{equation}
    A(\bm{X}) = \bm{W}_{up}\sigma(\bm{W}_{down}\bm{X}),
\end{equation}
where $\bm{W}_{down} \in \mathbb{R}^{C \times d}$ projects the feature into a lower-dimensional bottleneck space, $\bm{W}_{up} \in \mathbb{R}^{d \times C}$ restores the original dimension, $d \ll C$, and $\sigma$ is a non-linear activation function.

The adapted feature is obtained through residual refinement:
\begin{equation}
    \tilde{\bm{X}} = \bm{X} + \lambda A(\bm{X}),
\end{equation}
where $\lambda$ controls the contribution of the adapter branch.

\subsection{MoE-based Feature Adapter}

A single adapter applies one fixed transformation to all angiographic patterns. To improve flexibility, we replace it with a MoE-based feature adapter containing $M$ lightweight adapter experts:
\begin{equation}
    \{A_1, A_2, ..., A_M\}.
\end{equation}

The router predicts expert weights from the input feature:
\begin{equation}
    \bm{g} = \mathrm{Softmax}(\bm{W}_r \mathrm{Pool}(\bm{X})),
\end{equation}
where $\mathrm{Pool}(\cdot)$ aggregates token features and $W_r$ is a learnable projection. The output $\bm{g} \in \mathbb{R}^{M}$ represents the importance of each expert.

To reduce computation and encourage expert specialisation, we use top-$k$ routing. Let $\mathcal{T}_k(\bm{g})$ denote the indices of the top-$k$ expert weights. The normalised routing weight is:
\begin{equation}
    \hat{g}_i =
    \frac{g_i}{\sum_{j \in \mathcal{T}_k(\bm{g})} g_j},
    \quad i \in \mathcal{T}_k(\bm{g}).
\end{equation}

The MoE adapter output is:
\begin{equation}
    A_{\mathrm{MoE}}(\bm{X}) =
    \sum_{i \in \mathcal{T}_k(\bm{g})}
    \hat{g}_i A_i(\bm{X}).
\end{equation}

The final adapted feature is:
\begin{equation}
    \tilde{\bm{X}} = \bm{X} + \lambda A_{\mathrm{MoE}}(\bm{X}).
\end{equation}

This design enables complementary expert refinements for heterogeneous vessel and background appearances.


\subsection{Multi-level Adapter Insertion}

Coronary arteries appear at different spatial scales, from thick proximal segments to thin distal branches. We therefore insert MoE adapters into multiple transformer stages. Let $\{\bm{X}_1, \bm{X}_2, ..., \bm{X}_L\}$ denote encoder features from different stages. The adapted feature at stage $l$ is:

\begin{equation}
    \tilde{\bm{X}}_l =
    \begin{cases}
    \bm{X}_l + \lambda_l A_{\mathrm{MoE}}^{l}(\bm{X}_l), & l \in \mathcal{S}, \\
    \bm{X}_l, & l \notin \mathcal{S},
    \end{cases}
\end{equation}

where $\mathcal{S}$ denotes the selected adapter insertion stages. The adapted multi-level features are then passed to the decoder for binary mask prediction.

\subsection{Training Objective}

The model is trained using a combination of binary Dice loss and binary cross-entropy loss:

\begin{equation}
    \mathcal{L} = \mathcal{L}_{Dice} + \alpha \mathcal{L}_{BCE},
\end{equation}

where $\alpha$ balances the contribution of the BCE term, dice loss encourages foreground overlap, while BCE provides pixel-wise supervision for vessel and background classification.

\section{Experiments}

\subsection{Datasets}

\subsubsection{MOSXAV.} We use MOSXAV~\cite{xi2026diffusion,xi2026fsvos} as the main training and testing dataset. MOSXAV is a benchmark dataset for X-ray angiography video segmentation. In this work, we formulate the task as binary coronary artery segmentation. Specifically, the vessel annotation is used as foreground, and all remaining pixels are treated as background. All models are trained on the MOSXAV training set, selected using the validation set, and evaluated on the MOSXAV test set.

\subsubsection{XACV.} To evaluate external generalisation, we test the trained models on the XACV~\cite{xacv} dataset. XACV is an X-ray coronary angiography video dataset with manually annotated vessel segmentation ground truth. For XACV, we directly use the available binary vessel segmentation ground truth for external evaluation. In other words, the models trained on MOSXAV are directly evaluated on XACV to assess cross-dataset robustness.

\subsection{Implementation Details}

All input frames are resized to 1024$\times$1024. Pixel intensities are normalised to [-1,1]. The models are trained using AdamW with an initial learning rate of 3e-4 and a weight decay of 0.01. Training is performed for 40 epochs. The best model is selected according to the validation Dice score.

For the MoE-based feature adapter, we use $M=4$ experts and activate top-$k=2$ experts for each feature. The adapter bottleneck ratio is set to 1. MoE adapters are inserted into all ViT stages. The loss weight $\alpha$ is set to 0.7.

We compare the proposed method with several representative baselines, including U-Net~\cite{unet}, Attention U-Net~\cite{attentionunet}, nnU-Net~\cite{nnunet}, nnWNet~\cite{nnwnet}, AdapterSeg~\cite{adaptformer}, MaskVSC~\cite{maskvsc}, and a recently validated pseudo-label-based diffusion segmentation model for binary vessel segmentation~\cite{xi2026diffusion}.

\subsection{Evaluation Metrics}

We evaluate binary segmentation performance using Dice score, Intersection over Union (IoU), precision, and recall. Dice and IoU measure mask overlap, while precision and recall assess false-positive suppression and vessel recovery, respectively. Since F1 is equivalent to Dice for binary foreground segmentation when computed from pixel-level precision and recall, we omit F1 to avoid redundancy.

\section{Results}

\subsection{Results on MOSXAV}

Table~\ref{tab:mosxav} reports the quantitative results on MOSXAV. On the validation set, the proposed method achieves the best Dice, IoU, and recall, indicating improved vessel recovery and overall overlap performance. Although nnU-Net and nnWNet obtain higher precision, their lower recall suggests a more conservative prediction tendency. In contrast, our method provides a better precision-recall balance.

\begin{table*}[t]
\centering
\caption{Binary coronary artery segmentation results on the MOSXAV validation and test sets.}
\label{tab:mosxav}
\begin{tabular}{r|cccccccc}
\toprule
\multirow{2}{*}{Method~~~~~~~~~}
& \multicolumn{4}{c}{Validation Set}
& \multicolumn{4}{c}{Test Set} \\
\cmidrule(lr){2-5} \cmidrule(lr){6-9}
& Dice $\uparrow$ & IoU $\uparrow$ & Precision $\uparrow$ & Recall $\uparrow$
& Dice $\uparrow$ & IoU $\uparrow$ & Precision $\uparrow$ & Recall $\uparrow$ \\
\midrule
U-Net~\cite{unet} & 81.15 & 69.02 & 84.10 & 79.98 & 35.55 & 23.22 & 26.55 & 71.87 \\
Attention U-Net~\cite{attentionunet} & 81.76 & 69.86 & 86.50 & 79.11 & 38.68 & 25.85 & 29.43 & 70.27 \\
nnU-Net~\cite{nnunet} & 83.14 & 72.03 & 92.01 & 73.58 & 45.76 & 28.59 & 40.33 & 73.04 \\
Pseudo Diffusion~\cite{xi2026diffusion} & 80.82 & 66.89 & 82.36 & 72.63 & 34.78 & 22.97 & 24.26 & 68.11 \\
AdapterSeg~\cite{adaptformer} & 83.83 & 72.82 & 85.37 & 83.53 & 50.57 & 36.13 & 40.60 & \textbf{81.96} \\
MaskVSC~\cite{maskvsc} & 78.70 & 65.77 & 81.40 & 78.15 & 35.56 & 23.19 & 24.80 & 72.40 \\
nnWNet~\cite{nnwnet} & 84.07 & 72.99 & \textbf{93.03} & 74.62 & 46.18 & 29.17 & 41.26 & 74.29 \\
\textbf{Ours} & \textbf{84.75} & \textbf{74.05} & 85.22 & \textbf{85.97} & \textbf{53.47} & \textbf{39.09} & \textbf{46.29} & 77.74 \\
\bottomrule
\end{tabular}
\end{table*}

On the test set, our method achieves the best Dice and IoU, outperforming the strongest baseline, AdapterSeg. It also obtains the highest precision, suggesting improved suppression of vessel-like background responses. While AdapterSeg achieves the highest recall, its lower precision indicates more false-positive predictions. Overall, these results demonstrate that MoE-based feature adaptation improves segmentation robustness under challenging angiographic appearances.

\subsection{Results on XACV}

Table~\ref{tab:xacv} reports external testing results on XACV, where all models are trained on MOSXAV and evaluated without fine-tuning. The proposed method achieves the best Dice, IoU, and recall, showing stronger cross-dataset generalisation than the compared baselines. Although nnWNet obtains slightly higher precision, its lower recall indicates a more conservative prediction behaviour. In contrast, our method maintains competitive precision while recovering more vessel pixels, leading to the best overall overlap performance.

\begin{table}[t]
\centering
\caption{External binary vessel segmentation results on XACV.}
\label{tab:xacv}
\begin{tabular}{r|cccc}
\toprule
Method~~~~~~~~~ & Dice $\uparrow$ & IoU $\uparrow$ & Precision $\uparrow$ & Recall $\uparrow$ \\
\midrule
U-Net~\cite{unet} & 65.58 & 49.62 & 69.32 & 62.99 \\
Attention U-Net~\cite{attentionunet} & 64.30 & 48.25 & 71.28& 59.44 \\
nnU-Net~\cite{nnunet} & 67.02 & 51.23 & 77.61 & 59.66 \\
Pseudo Diffusion~\cite{xi2026diffusion} & 62.37 & 45.81 & 69.58 & 59.33 \\
AdapterSeg~\cite{adaptformer} & 68.41 & 52.76 & 76.20 & 62.95 \\
MaskVSC~\cite{maskvsc} & 60.56 & 44.37 & 69.20 & 54.65 \\
nnWNet~\cite{nnwnet} & 68.22 & 52.03 & \textbf{77.91} & 60.94 \\
\textbf{Ours} & \textbf{70.25} & \textbf{54.95} & 76.84 & \textbf{65.37} \\
\bottomrule
\end{tabular}
\end{table}

These results suggest that the proposed MoE-based adapter does not merely overfit to MOSXAV but improves robustness under unseen imaging distributions, where vessel contrast, background structures, and acquisition characteristics may differ from the training data.

\subsection{Ablation Study}

Table~\ref{tab:ablation} presents the ablation study on MOSXAV. The baseline backbone performs poorly without adapter-based adaptation, while adding a single adapter substantially improves Dice and IoU, confirming the importance of parameter-efficient feature adaptation. Replacing the single adapter with the proposed MoE adapter further improves performance, indicating that multiple expert pathways provide complementary refinements for heterogeneous angiographic appearances. 

\begin{table}[t]
\centering
\caption{Ablation study on MOSXAV.}
\label{tab:ablation}
\begin{tabular}{l|ccccc}
\toprule
~~~~~~~~~~~~Variant & Dice $\uparrow$ & IoU $\uparrow$ & Precision $\uparrow$ & Recall $\uparrow$ & Params \\
\midrule
Baseline backbone & 48.37 & 32.77 & 62.19 & 60.72 & 308M \\
+ Single adapter & 83.83 & 72.82 & 85.37 & 83.53 & 325M \\
+ MoE adapter, all experts & 84.25 & 73.91 & 85.08 & 84.88 & 356M \\
+ MoE adapter, top-1 & 83.76 & 73.19 & 85.13 & 83.47 & 356M \\
\textbf{+ MoE adapter, top-k} & \textbf{84.75} & \textbf{74.05} & \textbf{85.22} & \textbf{85.97} & 356M \\
\bottomrule
\end{tabular}
\end{table}

Among different routing strategies, top-$k$ routing achieves the best Dice, IoU, precision, and recall. Compared with top-1 routing or activating all experts, top-$k$ routing provides a better balance between expert specialisation and feature diversity.

\section{Conclusion}

We presented a MoE-based feature adapter framework for prompt-free binary coronary artery segmentation in X-ray angiography videos. The proposed method extends AdaptFormer-style lightweight adapters by introducing multiple expert branches and input-dependent top-$k$ routing, enabling adaptive feature refinement for heterogeneous angiographic appearances.

Experiments on MOSXAV and external testing on XACV demonstrate improved segmentation performance and stronger cross-dataset generalisation compared with representative baselines. These results indicate that MoE-based adapter learning is a promising strategy for robust coronary artery segmentation. Future work will explore temporal modelling across angiography videos and evaluate its impact on downstream clinical measurements such as vessel diameter estimation and stenosis quantification.

\section*{Acknowledgements}

This work was supported by EPSRC UK (EP/X023826/1).

\section*{Disclosure of Interests}

The authors have no competing interests to declare that are relevant to the content of this article.

\bibliographystyle{unsrt}  
\bibliography{references}

\end{document}